\documentclass[sigconf]{acmart}

\usepackage{algorithmic}
\usepackage{multicol}  
\usepackage{multirow}
\usepackage{tabularx}
\usepackage{booktabs}
\usepackage{subcaption}
\usepackage{amsmath}
\usepackage{caption}
\usepackage[linesnumbered, ruled]{algorithm2e}
\hypersetup{draft}
\usepackage{enumitem}
\usepackage{bbm}
\usepackage[export]{adjustbox}

\newcolumntype{C}{>{\centering\arraybackslash}X}
\newcolumntype{P}[1]{>{\centering\arraybackslash}p{#1}}

\newtheorem{problem}{Problem}

\newcommand{{\method}}{PxGNN}

\begin{document}
\fancyhead{}
\title{Towards Prototype-Based Self-Explainable Graph Neural Network}

\author{Enyan Dai}
\affiliation{The Pennsylvania State University}
\email{emd5759@psu.edu}

\author{Suhang Wang}
\affiliation{The Pennsylvania State University}
\email{szw494@psu.edu}

\begin{abstract}
Graph Neural Networks (GNNs) have shown great ability in modeling graph-structured data for various domains. However, GNNs are known as black-box models that lack interpretability. Without understanding their inner working, we cannot fully trust them, which largely limits their adoption in high-stake scenarios. Though some initial efforts have been taken to interpret the predictions of GNNs, they mainly focus on providing post-hoc explanations using an additional explainer, which could misrepresent the true inner working mechanism of the target GNN. The works on self-explainable GNNs are rather limited. Therefore, we study a novel problem of learning prototype-based self-explainable GNNs that can simultaneously give accurate predictions and prototype-based explanations on predictions. We design a framework which can learn  prototype graphs that capture representative patterns of each class as class-level explanations. The learned prototypes are also used to simultaneously make prediction for for a test instance and provide instance-level explanation. Extensive experiments on real-world and synthetic datasets show the effectiveness of the proposed framework for both prediction accuracy and explanation quality.
\end{abstract}

\maketitle
\section{Introduction}
Graph structured data such as traffic networks, social networks, and molecular graphs are very pervasive in real world. To model the graph structured data for various applications such as drug discovery~\cite{jiang2021could}, financial analysis~\cite{wang2019semi}, and recommendation system~\cite{wang2019knowledge}, various graph neural networks (GNNs)~\cite{bruna2013spectral,kipf2016semi,hamilton2017inductive} have been proposed and made remarkable achievements. The success of GNNs relies on the message-passing mechanism, i.e., the node representations in GNNs will aggregate the information from the neighbors to capture the attribute and topology information. Many message-passing mechanisms have been investigated to learn powerful representations from graphs, facilitating various tasks such as node classification~\cite{kipf2016semi,chen2020simple} and graph classification~\cite{xu2018powerful}. 

Despite the great success of GNNs in modeling graphs, GNNs have the same issue of lacking explainability as other deep learning models due to the high non-linearity in the model. In addition, the message-passing mechanism of GNNs that aggregates neighborhood features to capture topology information makes it more challenging to understand the predictions. The lacking of explainability in GNNs will  largely limit their adoption in critical applications  pertaining to fairness, privacy and safety. For instance, a GNN model may be trained to explore the proprieties of various drugs. However, due to the black-box characteristic of GNNs, it is unknown whether the rules learned by GNN model is consistent with the chemical rules in the real-world, which raises the concern of applying predictions that may threaten the drug safety.

Extensive approaches~\cite{shu2019defend,papernot2018deep} have been investigated to explain trained neural networks or give self-explainable predictions on independent and identically distributed (i.i.d) data such as images. However, they fail to generalize to GNNs due to the utilization of message-passing mechanism designed for relational information preservation.
Recently, some initial efforts~\cite{ying2019gnnexplainer,luo2020parameterized,huang2020graphlime,yuan2020xgnn,subgraphx_icml21} have been taken to address the explainability issue of GNNs. For example, GNNExplainer~\cite{ying2019gnnexplainer} explains the prediction of an instance by identifying the crucial subgraph of the instance's local graph. 
Model-level explanation is also investigated by generating graph patterns that maximize the prediction of each class by the target model~\cite{yuan2020xgnn}.
However, most of existing GNN explainers focus on the post-hoc explanations, i.e., learning an additional explainer to explain the predictions of a trained GNN.
Since the learned explainer cannot have perfect fidelity to the original model, the post-hoc explanations may misrepresent the true explanations of the GNNs~\cite{rudin2019stop}. 
Therefore, it is crucial to develop a self-explainable GNN, which can simultaneously give predictions and explanations.

\begin{figure}
    \centering
    \includegraphics[width=0.9\linewidth]{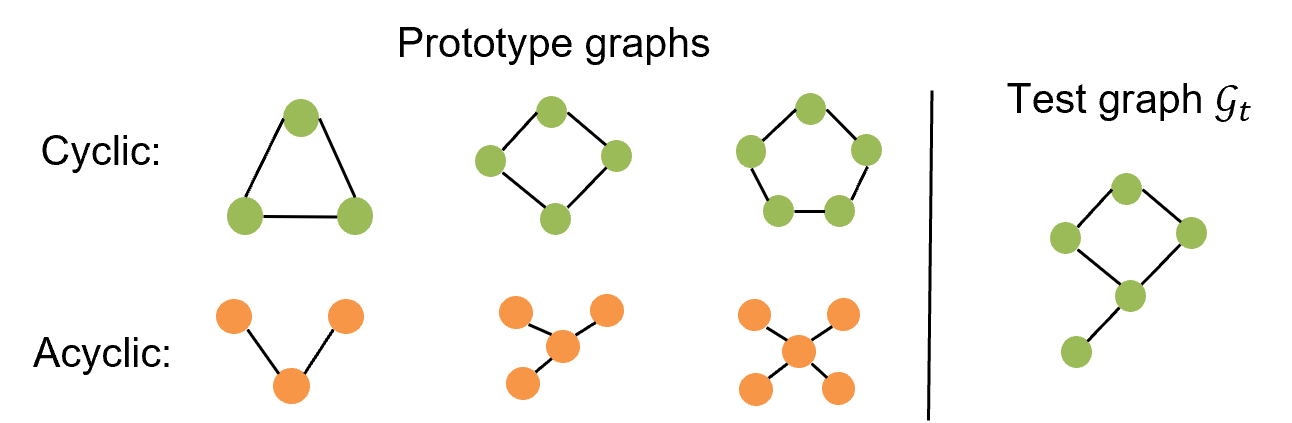}
    \vskip -1em
    \caption{An illustration of self-explanation with prototypes on classifying whether the test graph is cyclic.}
    \vskip -1.5em
    \label{fig:intro}
\end{figure}
One promising direction of self-explainable GNN is to learn prototype graphs of each class to present the  key patterns of each class and simultaneously conduct prediction and give explanations with the learned prototypes. The prototypes can provide class-level and instance-level self-explanations. Figure~\ref{fig:intro} gives an illustration of the prototype-based self-explanations on a toy classification problem. As shown in the figure, the problem is to predict whether a graph is cyclic or not. Taking the class cyclic as an example, the learned prototype graphs are typical patterns of cyclic graphs of various sizes. Therefore, the learned prototype graphs of class cyclic can provide class-level explanation to show representative graphs of class cyclic. For a test graph $\mathcal{G}_t$, we will match it with the prototype graphs of each class to give the prediction. Specifically, the instance-level explanation for predicting $\mathcal{G}_t$ can be: ``Graph $\mathcal{G}_t$ is classified as cyclic, because it is most similar to the second prototype graph of class cyclic."
Though promising, learning representative graphs for self-explainable classification remains an open problem.

Therefore, in this work,we investigate a novel problem of learning prototype-based self-explainable graph neural network. However, this is a non-trivial task. There are two main challenges: (\textbf{i}) how to efficiently learn high-quality prototypes that are representatives of each class for class-level explanation. Though some existing works~\cite{chen2019looks,li2018deep} have studied prototype learning for self-explanations, they are mainly proposed for i.i.d data. Recently, ProtGNN~\cite{zhang2021protgnn} applies a Monte Carlo tree search to identify subgraphs from raw graph as prototypes. However, the search algorithm is very time consuming. And the prototypes are limited to the subgraphs in the dataset, which might not be that representative; and (\textbf{ii}) how to simultaneously give an accurate prediction and provide correct prototype-based instance-level explanation. Different from images, the matching process between the test graph and prototype graphs cannot directly use simple metric such as Euclidean distance. Moreover, the supervision of the matching result is not available. How to effectively leverage the classification supervision for prototype learning and correct explanations needs further investigation.

In an attempt to address the above challenges, we develop a novel \underline{P}rototype-Based Self-E\underline{x}plainable \underline{GNN} ({\method})\footnote{Code and datasets will be released upon acceptance}. To efficiently obtain the prototype graphs, {\method} adopts a prototype graph generator to attain the prototype graphs from the learnable prototype embeddings.  A constraint on the learnable prototype embeddings and self-supervision from graph reconstruction are utilized to guarantee the quality of learned prototype embeddings and generated prototype graphs, respectively. An encoder is deployed to match the test graph with the generated prototype graphs for self-explainable classification. Since representative prototype graphs of a certain class is supposed to be similar to the test graphs in the same class, the labels can provide implicit supervision to ensure the representativeness of the prototype graphs and guide the matching process. More specifically, a novel classification loss is proposed to simultaneously ensure the accuracy of prediction and the quality of prototype-based instance-level explanation. And the classification loss is utilized to jointly train the  model and prototype embeddings to learn prototypes well represent their corresponding classes.   
In summary, our main contributions are:
\begin{itemize}[leftmargin=*]
    \item We investigate a novel problem of learning prototype graphs for self-explainable classification on graph-structured data;
    \item We develop a new framework {\method}, which learns an effective prototype generator with self-supervision to obtain high-quality prototype graphs for accurate predictions and explanations;
    \item We construct a synthetic dataset which can quantitatively evaluate the prototype-base explanation; and
    \item Extensive experiments on both real-world and synthetic datasets demonstrate the effectiveness of our {\method} in learning representative prototypes for accurate self-explainable classification.
\end{itemize}
\section{Related Work}

\subsection{Graph Neural Networks}
Graph Neural Networks (GNNs)~\cite{kipf2016semi,velivckovic2017graph,ying2018graph,bongini2021molecular} have shown great ability for representation learning on graphs, which facilitate various applications such as traffic analysis~\cite{zhao2020semi}, recommendation system~\cite{ying2018graph}, and drug generation~\cite{bongini2021molecular}. 
Generally, existing GNNs~\cite{kipf2016semi,levie2018cayleynets,velivckovic2017graph,xu2018representation,hamilton2017inductive,chen2018fastgcn,chiang2019cluster,chen2020simple} utilize a message-passing mechanism that a node's representation is updated by aggregating and combining the features from its neighbors. For example, GCN~\cite{kipf2016semi} averages the representations of neighbors and the target node followed by an non-linear transformation. 
GAT~\cite{velivckovic2017graph} adopts an attention mechanism to better aggregate the representations of the nodes from the neighbors. 
Recently, various GNN models are proposed to further improve the performance of GNNs~\cite{chen2018fastgcn,chen2020simple, li2019deepgcns,kim2021find,zhu2020self,qiu2020gcc,you2020graph}. For instance, FastGCN~\cite{chen2018fastgcn} is proposed to alleviate the scalability issue of GCN. In addition, some methods~\cite{chen2020simple, li2019deepgcns} focus on overcoming the oversmoothing issue of GCN and design deep GNNs to 
incorporate more hops of neighbors. Moreover, 
to facilitate the downstream tasks that are short of labels, self-supervised GNNs~\cite{kim2021find,zhu2020self,qiu2020gcc,you2020graph} are investigated to learn better representations.

\subsection{Explainability of Graph Neural Networks}
Despite the great success of graph neural networks, the problem of lacking explainability hinders the adoption of GNNs to various high-stake domains such as credit estimation. Though extensive methods~\cite{alvarez2018towards,hind2019ted,zeiler2014visualizing,yuan2019interpreting,du2018towards,selvaraju2017grad} have been proposed to explain neural networks, they are overwhelmingly developed for i.i.d data such as images and texts and cannot be directly applied to explain GNN models. 
Recently, some works in explainability of GNNs are emerging~\cite{ying2019gnnexplainer,luo2020parameterized,yuan2020xgnn,pope2019explainability,baldassarre2019explainability,dai2021towards}. 
The majority of these GNN explainers give the explanations by extracting the crucial nodes, edges, and/or node features. For instance, GNNExplainer~\cite{ying2019gnnexplainer} 
learns soft masks for edges and node features to  explain the predictions with the identified subgraphs and features. PGExplainer~\cite{luo2020parameterized} proposes to combine the global view of GNNs to facilitate the extraction of important graphs by applying a parameterized explainer.  
XGNN~\cite{yuan2020xgnn} generates representative graphs for a class as model-level explanations for graph classification. 

However, the aforementioned methods focus on post-hoc explanations for a trained GNN, i.e., they usually require additional explainer to explain the target GNN, which might misrepresent the decision reasons of the model. There are very few initial efforts for self-explainable GNNs~\cite{dai2021towards,zhang2021protgnn}, which aims to simultaneously give predictions and explanations on the predictions. SE-GNN~\cite{dai2021towards} simultaneously give the predictions and explanations of a target node by identifying the K-nearest labeled nodes. ProtGNN~\cite{zhang2021protgnn} is the most similar work to ours, which finds subgraphs from the raw graphs as prototypes to give self-explanations. However, ProtGNN only focuses on graph classification and the computational cost is very large due to the searching phase in finding the prototype subgraphs. 
Our proposed method is inherently different from this work: (\textbf{i}) we propose a novel prototype-based self-explainable GNN that is effective in both node and graph-level classification tasks; (\textbf{ii}) a prototype generator is deployed to efficiently learn more representative prototypes for self-explainable classification.
\section{PROBLEM DEFINITION}
We denote an attributed graph by $\mathcal{G}=(\mathcal{V},\mathcal{E}, \mathbf{X})$, where $\mathcal{V}=\{v_1,...,v_N\}$ is the set of $N$ nodes, $\mathcal{E} \subseteq \mathcal{V} \times \mathcal{V}$ is the set of edges, and $\mathbf{X}=\{\mathbf{x}_1,...,\mathbf{x}_N\}$ is the set of node attributes with $\mathbf{x}_i$ being the node attributes of node $v_i$. Let $\mathbf{A} \in \mathbb{R}^{N \times N}$ be the adjacency matrix of the graph $\mathcal{G}$, where $\mathbf{A}_{ij}=1$ if nodes ${v}_i$ and ${v}_j$ are connected; otherwise $\mathbf{A}_{ij}=0$. 
In this paper, we focus on both graph classification and node classification tasks. 
For graph classification, a set of labeled graphs $\mathcal{D}_L = \{\mathcal{G}_{i}, y_{i}\}_{i=1}^{|\mathcal{D}_L|}$ is given, where $|\mathcal{D}_L|$ is the size of the training set and $y_{i} \in \{1,\dots,C\}$ denotes the label of graph $\mathcal{G}_{i}$. We aim to give accurate prediction on unlabeled test set $\mathcal{D}_{U}=\{\mathcal{G}_u\}_{u=1}^{|\mathcal{D}_U|}$.  
As for semi-supervised node classification in a GNN with $n$ layers, the result is computed on the local graph $\mathcal{G}_{v}^{(n)}$ that contains $n$-hop neighbors of the center node $v$. Thus, the node classification task can be viewed as a special case of graph classification on local graph of the center node. The labeled set and unlabeled set of node classification can be written as $\mathcal{D}_L = \{\mathcal{G}_{v}^{(n)}:v \in \mathcal{V}_L\}$ and  $\mathcal{D}_U = \{\mathcal{G}_{v}^{(n)}:v \in \mathcal{V}_U\}$, where $\mathcal{V}_L$ and $\mathcal{V}_U$ are the sets of labeled nodes and unlabeled nodes in the graph, respectively. 

In this paper, we aim to develop a self-explainable GNN that can accurately predict labels and provide both class-level and instance-level explanations. \textit{First}, for class-level explanation, for each class $l \in \{1, \dots, C\}$, we will  learn $K$ prototype graphs $\mathcal{P}_{l}=\{\mathcal{\Tilde{G}}_{li}\}_{i=1}^K$, which are representative prototypical graph patterns of class $l$. \textit{Second}, for a test graph $\mathcal{G}_t$, we will match the test graph with the learned prototype graphs and predict the label based on prototype graphs. Then, the instance-level explanation can be: ``Graph $\mathcal{G}_t$ is predicted as class $l$, because it is  similar with the prototype graphs of class $l$." 
With the description above, the problem of learning prototype-based self-explainable GNN can be written as:
\begin{problem}
Given the dataset $\mathcal{D}$, where $\mathcal{D} = \mathcal{D}_L$ for graph classification and  $\mathcal{D} = \mathcal{D}_L \cup \mathcal{D}_U$ for semi-supervised node classification, 
we aim to learn a self-explainable GNN $f: \mathcal{G} \rightarrow y$, which can generate prototype graphs $\{\mathcal{P}_l\}_{l=1}^C$ as class-level explanation and give accurate prediction to each unlabeled graph $\mathcal{G}_u \in \mathcal{D}_U$ along with the best matched prototype graphs $\mathcal{\Tilde{G}}_{u^*}$ as the instance-level explanation. 
\end{problem}

\section{Methodology}
\begin{figure*}
    \centering
    \includegraphics[width=0.90\linewidth]{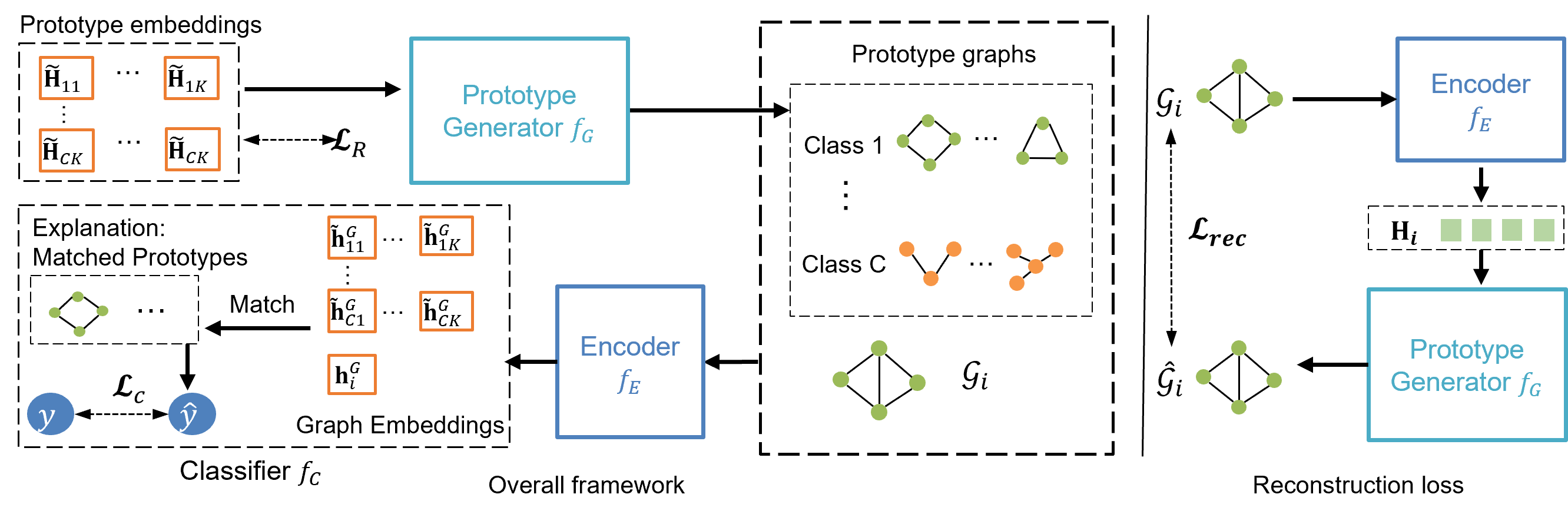}
    \vskip -1.5em
    \caption{The overall framework of our proposed {\method}.}
    \vskip -1em
    \label{fig:framework}
\end{figure*}

In this section, we present the details of the proposed framework {\method}. In particular, {\method} will generate prototype graphs that show the representative graph patterns of each class as class-level explanation. Meanwhile, for a test graph $\mathcal{G}_t$, the prediction will be provided based on the similarity scores between $\mathcal{G}_t$ and prototype graphs in different classes. Those most similar prototypes to $\mathcal{G}_t$ also serve as instance-level explanation. In essence, we are faced with two challenges: (i) how to design the framework to effectively learn high-quality prototype graphs that capture representative patterns; and (ii) how to ensure the accuracy of the classification on the test instance and the correctness of the corresponding instance-level explanation. 
To address the above challenges, we propose a novel framework {\method}, which is illustrated in Figure~\ref{fig:framework}. It is composed of a prototype generator $f_G$, an encoder $f_E$ and a prototype-based classifier $f_C$. 
The prototype generator $f_G$ takes the learnable prototype embeddings as input to generate prototype graphs. The encoder $f_E$ is adopted to match the test graph with the prototype graphs in feature space. Finally, based on the similarity scores between the test graph and prototype graphs in different classes, the classifier $f_C$ can give the prediction with self-explanation.
To generate high-quality prototype graphs, we utilize the self-supervision of the graph reconstruction to train the prototype generator and constrain the learnable prototype embedding.
In addition, a novel classification loss is proposed to ensure the accuracy of predictions and the quality of explanations. Note that our {\method} is flexible to both node classification and graph classification. The main difference between them is that the prototype graphs in node classification focus on the local graph of the test node instead of a whole graph of an instance.
Next, we will use graph classification as an example to introduce each component in detail. 

\subsection{Prototype Graph Learning}
\label{sec:4.1}
Our {\method} relies on realistic and representative prototype graphs to give accurate predictions and explanations. To obtain high-quality prototype graphs, we adopt two strategies for prototype graph generation: (i) self-supervision of reconstructing graphs from the embeddings encoded by $f_E$ is utilized to train the prototype-generator $f_G$; and (ii) we propose an effective method to initialize the prototype graphs. The initialized prototype graphs are further used to guide the learning of prototype embeddings and the prototype graph generation. 
\subsubsection{ Prototype Generator and Encoder}
To generate the prototype graph using the prototype generator, each prototype graph $\mathcal{\Tilde{G}}_{lk}$ is associated with a set of learnable prototype embeddings $\mathbf{\Tilde{H}}_{lk}=[\mathbf{\Tilde h}_1,\dots,\mathbf{\Tilde h}_{N_{lk}}]$, where $\mathbf{\Tilde h}_i$ denotes the embeddings of $i$-th node in $\mathcal{\Tilde{G}}_{lk}$ and $N_{lk}$ is the number of nodes in $\mathcal{\Tilde{G}}_{lk}$. 
These embeddings are learned together with other components of {\method} and initialized using an effective strategy to facilitate the learning process, which will be discussed in Sec.~\ref{sec:init}. $N_{lk}$ can be set according to the various sizes of the initialized prototype graphs to learn prototype graphs in different sizes.
The generator $f_{G}$ takes in $\mathbf{\Tilde{H}}_{lk}$ and generates $\mathcal{\Tilde{G}}_{lk}$. 
For an attributed graph, the prototype graph generator $f_{G}$ needs to generate both node features and graph topology. 
Given the node embedding $\mathbf{\Tilde h}_i$, the attributes of node $v_i$ is generated by a MLP as
\begin{equation}
    \mathbf{\tilde{x}}_i = \text{MLP}(\mathbf{\Tilde{h}}_i),
    \label{eq:attr_dec}
\end{equation}
For topology generation, we predict the link weight between node $v_i$ and $v_j$ as 
\begin{equation}\small
    \mathbf{\Tilde S}_{ij}=\sigma(\mathbf{W} \cdot \text{CONCAT} ( \mathbf{\tilde{x}}_i, \mathbf{\tilde{x}}_j)),
\end{equation}
where $\sigma$ is the sigmoid function and $\mathbf{W}$ is the learnable parameters. 

To ensure that the generated prototype graph patterns are reasonable, self-supervision from the dataset $\mathcal{D}$ is applied on prototype generator $f_G$ to generate realistic graphs that follow the distribution of $\mathcal{D}$. Following previous graph generative model~\cite{kipf2016variational}, graph reconstruction loss and an auxiliary encoder $f_E$ are deployed to train the graph generator. Specifically, the encoder $f_E$ takes the graph $\mathcal{G}_i \in \mathcal{D}$ as input and outputs the node embeddings $\mathbf{H}_i$. The encoder adopts GNN to obtain node embeddings that capture both node features and graph topology information, which can be written as
\begin{equation}\small
    \mathbf{H}_i = \text{GNN}(\mathbf{X}_i, \mathbf{A}_i),
    \label{eq:enc_node}
\end{equation}
where $\mathbf{X}_i$ and $\mathbf{A}_i$ are the node attribute matrix and adjacency matrix of graph $\mathcal{G}_i \in \mathcal{D}$. With the embedding matrix $\mathbf{H}_i$, we can get the reconstructed attribute matrix $\mathbf{\hat{X}}_i$ using Eq.(\ref{eq:attr_dec}). The reconstruction loss for attributes would be:
\begin{equation}
    \mathcal{L}_{rec}^{X} = \frac{1}{|\mathcal{D}|} \sum_{\mathcal{G}_i \in \mathcal{D}}\| \hat{\bf X}_i - \mathbf{X}_i \|^2.
    \label{eq:rec_attr}
\end{equation}
As for the reconstruction loss on adjacency matrix, a negative sampling strategy~\cite{mikolov2013distributed} is applied to avoid the domination of zero entries in adjacency matrix.  Specifically, for node $v_j^i$ in graph $\mathcal{G}_i$, we randomly select $Q$ nodes from $\mathcal{G}_i$ that are not connected with $v_j^i$ as negative samples in graph classification. For node classification, the negative samples are obtained from the whole graph instead of the computation graph $\mathcal{G}_i$.
The loss function can be formulated as:
\begin{equation}\small
\begin{aligned}
    \mathcal{L}_{rec}^{A} = \frac{1}{|\mathcal{D}|}\sum_{\mathcal{G}_i \in \mathcal{D}} \sum_{v_j^i \in \mathcal{G}_i} \sum_{v_k^i \in \mathcal{N}(v_j^i)} \big[-\log(\mathbf{S}_{jk}^i) - \sum_{n=1}^Q \mathbb{E}_{v_n^i \sim P_n(v_j^i)} \log(1-\mathbf{S}_{jn}^i)\big],
    \label{eq:rec_adj}
\end{aligned}
\end{equation}
where $\mathcal{N}(v_j^i)$ denotes the neighbors of node $v_j^i$, and $P_n(v_j^i)$ is the distribution of negative samples of node $v_j^i$. $\mathbf{S}_{jk}^i$ is the predicted link weight between $v_j^i$ and $v_k^i$. Combining Eq.(\ref{eq:rec_attr}) and Eq.(\ref{eq:rec_adj}), the overall reconstruction loss can be written as:
\begin{equation}\small
    \min_{\theta_G,\theta_E} \mathcal{L}_{rec} = \mathcal{L}_{rec}^{X} + \mathcal{L}_{rec}^{A}.
    \label{eq:rec}
\end{equation}
where $\theta_G$ and $\theta_E$ are the parameters of prototype generator $f_G$ and encoder $f_E$. With Eq.(\ref{eq:rec}), the prototype generator can generate realistic prototype graphs given well learned prototype embeddings.

\subsubsection{Initialization and Constraint on Prototype Embeddings} \label{sec:init}
With Eq.(\ref{eq:rec}), we can get an effective prototype generator. 
However, to obtain useful prototype graphs, there are two requirements for prototype embeddings: (i) to make sure that the generated prototypes are realistic, the prototype embeddings should be in the same latent space as that learned by the encoder because the generator is trained to generate realistic graph when the inputs are from that space; and (ii) to make sure that the generated prototypes are representative of each class, the prototype embeddings should be centroids of the latent space learned by encoder. Therefore, we propose to firstly pretrain the encoder. Then, we initialize the prototype embeddings by identifying the representative graphs whose graph embeddings are in the center of the learned latent space. Specifically, the encoder is pretrained by the labeled 
graphs to learn the latent space that  carries the embedding information of the graph. This label prediction of $\mathcal{G}_i \in \mathcal{D}_L$ with the encoder $f_E$ can be obtained by
\begin{equation}\small
    \mathbf{h}_i^G = \text{READOUT}(f_E(\mathbf{A}_i,\mathbf{X}_i)), \quad
    \hat{y}_i = \text{softmax}(\mathbf{W}_{C} \cdot \mathbf{h}_i^G),
    \label{eq:graph_enc}
\end{equation}
where $\mathbf{h}_i^G$ is the graph embedding of $\mathcal{G}_i$, $\text{READOUT}$ is flexible to max pooling or mean pooling, and $\mathbf{W}_C$ is the learnable weigh matrix for classification. $\mathbf{A}_i$ and $\mathbf{X}_i$ are the adjacency matrix and feature matrix of $\mathcal{G}_i$. We then pretrain encoder and decoder as:
\begin{equation}\small
    \min_{\theta_E,\theta_G,\mathbf{W}_C} \frac{1}{|\mathcal{D}_L|}\sum_{\mathcal{G}_i \in \mathcal{D}_L} l(\hat{y}_i, y_i) + \alpha \mathcal{L}_{rec},
    \label{eq:pretrain}
\end{equation}
where $l(\cdot)$ denotes the cross entropy loss, and $y_i$ is the label of $\mathcal{G}_i$. With the pretrained encoder, we can find the representative graphs in the center of latent space. For each class $l \in \{1,\dots,C\}$, we will apply K-Means to cluster $\{\mathbf{h}_{i}^G: \hat{y}_i=l, \mathcal{G}_i \in \mathcal{D}\}$, i.e., the embeddings of graphs that are predicted as class $l$. The number of clusters/centroids of each class are pre-defined as $K$. Let $\mathbf{h}_{lk}^C$ denote the centroid of $k$-th cluster of class $l$,
we can select the graph whose graph embedding is closest to $\mathbf{h}_{lk}^C$ by
\begin{equation}\small
    \mathcal{G}_{lk}^{init} = \arg \min_{\mathcal{G}_i \in \mathcal{D}} \|\mathbf{h}_i^G - \mathbf{h}_{lk}^C\|^2.
    \label{eq:get_init}
\end{equation}
Then, $\mathbf{H}^{init}_{lk}$, i.e., the initialization of prototype embedding for prototype graph $\mathcal{\Tilde{G}}_{lk}$, can be attained as $\mathbf{H}^{init}_{lk}=f_E(\mathcal{\Tilde{G}}_{lk}^{init})$. Since the optimal prototype graphs should not differ a lot from the initialization graphs and to make the prototype embeddings remains in the desired latent space, we further add a constraint to regularize the learning process of prototype embeddings as:
\begin{equation}\small
    \mathcal{L}_{R} = \frac{1}{C \cdot K} {\sum}_{l=1}^C {\sum}_{k=1}^K \|\mathbf{\Tilde H}_{lk}-\mathbf{H}_{lk}^{init}\|_F^2.
\end{equation}
\subsubsection{Prototype Graph Generation} With the prototype embeddings and generator, we can generate the prototype graphs. However, directly using the generated adjacency matrix would lead to a fully-connected graph which is difficult to interpret. Hence, we utilize the initialized graph $\mathcal{G}^{init}_{lk}$ to help remove unnecessary links to obtain a realistic sparse graph.
For a node pair $(v_i,v_t)$ that is not a link of $\mathcal{G}^{init}_{lk}$, it is unlikely to be linked in the learned prototype graphs. Hence, we will set a relatively high threshold for link generation.  
Considering that a link $(v_i,v_j)$ in $\mathcal{G}^{init}_{lk}$ is more likely to also appear in the learned prototype graph $\mathcal{\Tilde{G}}_{lk}$, a lower threshold of link elimination will be set. 
Let $\mathbf{\Tilde S}_{ij}$ denotes the predicted probability that node $v_i \in \mathcal{\Tilde{G}}_{lk}$ and $v_j \in \mathcal{\Tilde{G}}_{lk}$ are connected. 
The final adjacency matrix $\mathbf{\Tilde{A}}$ of the prototype graph $\mathcal{\Tilde{G}}_{lk}$ can be written as:
\begin{equation}\small
    \mathbf{\Tilde{A}}_{ij} = \left\{ \begin{array}{ll}
         \mathbf{\Tilde S}_{ij} & \mbox{if  $v_j \in \mathcal{N}^{init}(v_i) $ and $\mathbf{\Tilde S}_{ij} > T_l$} ; \\
         \mathbf{\Tilde S}_{ij} & \mbox{if $v_j \notin \mathcal{N}^{init}(v_i)$ and $\mathbf{\Tilde S}_{ij} > T_h$}\\
        0 & \mbox{else},\end{array} \right.
        \label{eq:generate_graph},
\end{equation}
where $1 \geq T_h \geq T_l \geq 0 $ and $\mathcal{N}^{init}(v_i)$ denotes the neighbor of $v_i \in \mathcal{G}_{lk}^{init}$.  As for the attributes of $\mathcal{\Tilde{G}}_{lk}$, it can be directly obtained by Eq.(\ref{eq:attr_dec}) with the learned prototype embeddings $\mathbf{\Tilde H}_{lk}$ as input. Note that domain specific constraints can also be incorporated in this process to generate more realistic prototype graphs.

\subsection{Self-Explainable Prediction with Prototypes}
With the prototype graph generation described in Section~\ref{sec:4.1}, we are able to conduct the prediction by finding the prototype graphs that are  similar to the test graph to give predictions and instance-level explanations. Next, we will present the details of the classification, self-explanation, and the loss function that facilitate both classification accuracy and explanation quality.  
\subsubsection{Prediction with Prototypes} Intuitively, if a test graph $\mathcal{G}_t$ is more similar with the prototype graphs in class $l$, the label of $\mathcal{G}_t$ is more likely to be class $l$. 
Following previous works in similarity metric learning~\cite{dai2021towards}, we use $f_E$ to learn graph representation followed by a similarity function. Let $\mathbf{h}_t^G$ and $\mathbf{\Tilde{h}}_{lk}^G$ represent the  graph embeddings of $\mathcal{G}_t \in \mathcal{D}$ and $\mathcal{\tilde{G}}_{lk} \in \mathcal{P}_l$ encoded by Eq.(\ref{eq:graph_enc}), the similarity score is calculated as:
\begin{equation}
    s(\mathcal{G}_t,\mathcal{\tilde{G}}_{lk})=sim(\mathbf{h}_t^G,\mathbf{\Tilde{h}}_{lk}^G),
    \label{eq:sim}
\end{equation}
where $sim$ is the similarity function which can be cosine similarity or distance-based similarity. With Eq.(\ref{eq:sim}), we can find the prototype graphs that are similar to the test graph $\mathcal{G}_t$ and predict the label with weighted average of the class of the nearest prototypes. Let $\mathcal{P}_t=\{\mathcal{\Tilde{G}}_1^t,\dots,\mathcal{\Tilde{G}}_M^t\}$ be the identified $M$-nearest prototype graphs. The weight $a_{ti}$ of the $i$-th nearest prototype graphs is computed as:
\begin{equation}\small
    a_{ti} = \frac{\exp(s(\mathcal{G}_t,\mathcal{\Tilde{G}}_i^t)/\tau)}{\sum_{i=1}^{M}\exp(s(\mathcal{G}_t,\mathcal{\Tilde{G}}_i^t)/\tau)},
\end{equation}
where $\tau$ is the temperature parameter. Let $\mathbf{y}_i^t$ denotes the one-hot class vector of prototype graph $\mathcal{\Tilde{G}}_i^t$, the final class distribution prediction of $\mathcal{G}_t$ is given as:
\begin{equation}\small
    \hat{\mathbf{y}}_t = {\sum}_{i=1}^M a_{ti}\cdot\mathbf{y}_i^t
    \label{eq:pred}
\end{equation}

\subsubsection{Explainability of \method} \textit{First}, the learned prototype graphs $\mathcal{P}_l$ is the class-level explanation, which shows the representative patterns of graph in class $l$. \textit{Second}, for the prediction of the test graph $\mathcal{G}_t$, the obtained similarity scores between $\mathcal{G}_t$ and the prototype graphs can explain the predicted label. Moreover, we can identify the prototype graph in class $\hat{y}_t$ that is most similar with $\mathcal{G}_t$ as the instance-level explanation by
\begin{equation}\small
   \mathcal{\tilde{G}}_{t}^* = \arg \max_{\mathcal{\Tilde{G}}_i \in \mathcal{P}_{\hat{y}_t} } s(\mathcal{G}_t, \mathcal{\Tilde{G}}_i).
\end{equation}
Since the prediction of $\mathcal{G}_t$ is based on several prototypes, the instance-level explanation can be flexible to be several most similar prototype graphs in class $\hat{y}_t$.

\subsubsection{Classification Loss} 
To ensure the accuracy of the proposed {\method}, we apply a classification loss to leverage the supervision from the labeled set $\mathcal{D}_L$. The major idea is that a labeled graph $\mathcal{G}_i \in \mathcal{D}_L$ whose label is $y_i$ should be more similar with the prototype graphs in $\mathcal{P}_{y_i}$ than other prototype graphs that do not belong to class $y_i$. Therefore, in the classification loss, the prototype graphs in $\mathcal{P}_{y_i}$ is set as positive samples; while the rest prototype graphs are set as negative samples. In addition, the prototype embeddings are jointly trained with the model parameters to update the prototype graphs to be more representative. The objective function of classification loss can be formally written as: 
\begin{equation}\small
    \min_{\theta_G,\theta_E,\mathcal{\Tilde {H}}} \mathcal{L}_c = \frac{1}{|\mathcal{D}_L|} \sum_{\mathcal{G}_i \in \mathcal{D}_L} -\log \frac{\sum_{\mathcal{\tilde G}_k \in \mathcal{P}_{y_i}} \exp (s(\mathcal{G}_i, \mathcal{\tilde G}_k)/\tau)}{\sum_{\mathcal{\tilde G}_k \in \mathcal{P}} \exp (s(\mathcal{G}_i, \mathcal{\tilde G}_{k})/\tau)},
    \label{eq:class}
\end{equation}
where $\tau$ is the temperature hyperparameter, and $\mathcal{\tilde H}$ is the set of prototype embeddings of all classes, and $\mathcal{P}$ represents all the prototype graphs. With Eq.(\ref{eq:class}), the similarity scores of a graph with prototype graphs in different classes will be minimized, and the similarity scores between a graph and the prototype graphs share the same class will be maximized. As a result, accurate predictions can be given. Moreover, it provides supervision to guide the similarity modeling to give correct instance-level explanation. 


\subsection{Final Objective Function of \method}
With the reconstruction loss for prototype learning, the constraint from the initialized prototype graph embeddings, and the classification that facilitate the prediction and explanation quality, the final objective function of {\method} is given as:
\begin{equation}
    \min_{\theta,\mathcal{\Tilde {H}}} \mathcal{L}_c + \alpha \mathcal{L}_{rec} + \beta \mathcal{L}_{R},
    \label{eq:final}
\end{equation}
where $\theta$ and $\mathcal{\Tilde {H}}$ denote all model parameters of {\method} and the set of learnable prototype embeddings, respectively. $\alpha$ and $\beta$ are hyperparameters to control the contribution of the reconstruction loss and constraint term on prototype embeddings, respectively. 

\section{experiments}
We conduct extensive experiments on both synthetic datasets and real-world graphs to answer the following research questions.
\begin{itemize}[leftmargin=*]
    \item \textbf{RQ1} Can our proposed {\method} learn high-quality prototypes for accurate predictions and explanations?
    \item \textbf{RQ2} How the number of prototypes will affect the accuracy and quality of the learned prototypes for explanations.
    \item \textbf{RQ3} How does each component of our proposed method affect the prediction accuracy and prototype quality?
\end{itemize}

\subsection{Datasets}
To evaluate our {\method} quantitatively and qualitatively, we conduct experiments on four real-world datasets and two synthetic datasets, which include node classification and graph class tasks.  The statistics of the datasets are presented in Table~\ref{tab:dataset}.

\subsubsection{Real-World Datasets}
We use two datasets for node classification, i.e., Cora and Pubmed~\cite{kipf2016semi}, and two datasets for graph classification, i.e., MUTAG~\cite{wu2018moleculenet} and Graph-SST2~\cite{yuan2020explainability}. \textbf{Cora} and \textbf{Pubmed} are both citation networks. We reduce the bag-of-words feature dimensionality with PCA to avoid the negative effects of the feature sparsity.  \textbf{MUTAG} is a dataset of molecule graphs labeled according to their mutagenic effect. And carbon ring with chemical groups $NO_2$ are known to be mutagenic, which can serve as the ground truth of MUTAG. As for \textbf{Graph-SST2}, it is a sentiment graph dataset for graph classification. The nodes and edges represent the words and their relations between each other. Each graph is labeled by its sentiment, which can be classified as positive or negative. The datasets splits  are the same as the cited papers. 

\subsubsection{Synthetic Datasets} 
We also conduct experiments on two synthetic datasets, i.e., BA-Shapes and Syn-SST, which provide the ground-truth prototype graphs for node classification and graph classification, respectively. 

\noindent \textbf{BA-Shapes}~\cite{ying2019gnnexplainer}: It is a single graph consisting of a base Barabasi-Albert (BA) graph attached with “house”-structured motifs. Nodes in the base graph are labeled with 0. Nodes at the top/middle/bottom of the “house” are labeled with 1,2,3, respectively. Therefore, the ``house"-structured motif is the ground-truth prototype graphs of class 1, 2, and 3. Following previous works~\cite{ying2019gnnexplainer,dai2021towards}, random edges are added to perturb the graph. Node degrees are assigned as node features. The dataset split is the same as the cited papers. 

\noindent \textbf{Syn-SST}: To obtain a dataset with ground-truth prototype graphs for graph classification, we sample graphs from Graph-SST2 as motifs to build Syn-SST. More specifically, we randomly sample five graphs for each class as the corresponding ground-truth prototypes. For each prototype graphs, we synthesize 20 perturbed versions by adding various levels of structural noises and randomly removing/injecting nodes. As a result, we obtain 100 graphs for each class in Syn-SST. For the dataset split, we randomly sample 0.5/0.25/0.25 molecules as train/validation/test set. The split sets have no overlap with each other.

\begin{table}[t]
    \small
    \caption{Statistics of datasets.}
    \vskip-1.5em
    \centering
    \begin{tabularx}{0.95\linewidth}{p{0.17\linewidth}CCCCC}
    \toprule
         & \#Nodes & \#Edges & \#Features & \#Classes & \#Graphs\\
    \midrule
    Cora & 2,708  & 5,429 & 1,433 & 7 & 1\\
    Pubmed & 19,171 & 44,338  & 500 & 3 & 1\\
    BA-Shapes & 700 & 4,421 & - & 4 & 1 \\
    \midrule
    MUTAG & 3,371 & 7,442 & 7 & 2 & 188 \\ 
    Graph-SST2 & 714,325 & 1,288,566 & 768 &2 & 70,042\\
    Syn-SST &  2,877 & 5,270 & 768 & 2 & 200\\
    \bottomrule
    \end{tabularx}
    \label{tab:dataset}
    \vskip -2em
\end{table}

\subsection{Experimental Settings}

\subsubsection{Compared Methods} 
To evaluate {\method}, we compare with the following state-of-art GNNs and self-explainable GNNs.
\begin{itemize}[leftmargin=*]
    \item \textbf{GCN}~\cite{kipf2016semi}: This is a spectral-based graph neural network, which aggregates the neighbor information by averaging their features. 
    \item \textbf{GIN}~\cite{xu2018powerful}: To increase the ability of capturing topology information in the graph, a MLP is applied in the aggregation phase.
    \item \textbf{ProtGNN}~\cite{zhang2021protgnn}: It employs a search algorithm to find the subgraphs as prototypes and adopts a prototype layer~\cite{chen2019looks} for self-explainable prediction. ProtGNN is designed for graph classification.  We extend it to node classification by giving predictions based on the local graphs of target nodes.
    \item \textbf{SE-GNN}~\cite{dai2021towards}: It focuses on node classification. The K-nearest labeled nodes' local graphs for each test node are identified for self-explainable predictions. For each class, we select the mostly selected labeled node' local graph as the prototype.
\end{itemize}
We also utilize the following post-hoc GNN explainers in extracting important subgraphs to get prototypes to compare with our proposed framework in prototype-based explanations.
\begin{itemize}[leftmargin=*]
    \item \textbf{GNNExplainer}~\cite{ying2019gnnexplainer}: It aims to find the subgraph that results similar predictions as the complete computation graph to give post-hoc explanations for a trained GNN.
    \item \textbf{PGExplainer}~\cite{luo2020parameterized}: It adopts a MLP-based explainer to obtain the important subgraphs from a global view to reduce the computation cost and obtain better explanations.
    \item \textbf{SubgraphX}~\cite{yuan2020explainability}: Instead of assuming that edges are independent to each other, SubgraphX explores different subgraphs by Monte Carlo tree search with Shapley values as importance measure.
\end{itemize}
Since they are post-hoc instance-level explainers, to generate prototypes, we first find representative graphs whose embedding lies in the center of embeddings in each class. We then obtain prototypes by extracting important subgraphs of the center graphs with the aforementioned post-hoc GNN explainers.

\subsubsection{Implementation Details} \label{sec:imple} For our {\method}, a two layer GCN is applied as the encoder. A two-layer MLP is deployed for attribute generation in the prototype generator. For the similarity function $s$ in the classifier, we apply an Euclidean distance-based similarity metric. All the hyperparameters are tuned based on the performance on validation set and the quality of learned prototype graphs. We vary $\alpha$ and $\beta$ as $\{100, 10, 1, 0.1, 0.01\}$ and $\{10,3,1,0.3,0.1\}$. As for the number of prototypes per class $K$, we search from $\{2,3,\dots,6\}$. We set $M$, i.e., 
the number of identified most similar prototypes for prediction, the same as $K$ for each dataset. For all experiments, $T_h$, $T_l$, $Q$ and $\tau$ are fixed as 0.8, 0.2, 50, and 1, respectively.  All the experiments are conducted 5 times. The hyperparameters of baselines are also tuned with grid search for fair comparisons.  

\begin{table}[t!]
    \caption{Classification accuracy (\%) on real-world datasets.}
    \vskip -1em
    \centering
    \small
    \resizebox{0.98\linewidth}{!}{
    \begin{tabular}{lccccc}
        \toprule
        Dataset & GCN & GIN & SE-GNN & ProtGNN &Ours  \\
        \midrule
        Cora & 80.8$\pm 1.2$ & 79.4 $\pm 1.2$  & 80.3 $\pm 0.7$& 75.2 $\pm 1.3$  & \textbf{82.0} $\pm 0.5$ \\
        Pubmed & 78.4 $\pm 0.4$ & 77.3 $\pm 0.6$ & 79.2 $\pm 0.7$ & 75.3 $\pm 1.2$ & \textbf{79.5} $\pm 1.2$\\
        \midrule
        MUTAG & 86.4 $\pm 1.8$ & 87.1 $\pm 1.0$ & - & 84.6 $\pm 1.0$ & \textbf{87.1} $\pm 1.3$   \\
        Graph-SST2 & 87.2 $\pm 0.3$ & 87.3 $\pm 0.2$ &- & 86.5 $\pm 2.4$&  \textbf{87.6} $\pm 0.5$ \\
        \bottomrule
    \end{tabular}
    }
    \vskip -1em
    \label{tab:pred}
\end{table}

\begin{table*}[t!]
    \centering
    \small
    \caption{Comparisons with baseline methods on the quality of obtained prototype graphs. }
    \vskip -1.5em
    \begin{tabularx}{0.98\linewidth}{|C|l|CCCCCC|}
    \hline
        Dataset& Metric & GNNexplainer & PGexplainer & SubgraphX &  SE-GNN & ProtGNN & Ours \\
        \hline
        \multirow{2}{*}{Cora}& Confidence Score (\%) & 89.4 $\pm 17.4$  & 70.9 $\pm 19.5$ & 41.3 $\pm 1.3$ & 93.3 $\pm 12.5$  & 41.7 $\pm 20.2$ & \textbf{98.4} $\pm 1.9$\\ 
        & Silhouette Score & 0.133 & 0.113 & 0.110 & 0.097 & 0.083  & \textbf{0.228}\\ 
        \hline
        \multirow{2}{*}{Pubmed}& Confidence Score (\%) & 93.3 $\pm 11.2 $ &89.2 $\pm 9.4$ & 72.3 $\pm 15.5$ &  94.8 $\pm 0.8$ & 66.7 $\pm 4.9$ &  \textbf{99.9} $\pm 0.1$\\ 
        & Silhouette Score & 0.264 & 0.205 & 0.123 & 0.210 & 0.239 &  \textbf{0.423}\\ 
        \hline
        \multirow{2}{*}{MUTAG}& Confidence Score (\%) & 76.4 $\pm 23.4$ & 70.5 $\pm 20.3$ & 70.5 $\pm 18.4 $ & - & 55.9 $\pm 14.5$ &  \textbf{94.5} $\pm 4.9$\\ 
        & Silhouette Score & 0.579 & 0.661 & 0.660 & - & 0.629 & \textbf{0.745}\\ 
        \hline
        \multirow{2}{*}{Graph-SST2}& Confidence Score (\%) & 87.7 $\pm 20.3$ & 90.2 $\pm 1.0$ & 89.1 $\pm 15.1$ & -& 62.3 $\pm 15.9$ & \textbf{94.9} $\pm 6.8$\\ 
        & Silhouette Score & 0.259 & 0.226 & 0.249 & - & 0.037 &  \textbf{0.309} \\ 
        \hline
    \end{tabularx}
    \label{tab:proto}
\end{table*}

\begin{table}[t!]
    \centering
    \small
    \caption{Distance from the ground-truth prototypes.}
    \vskip -1em
    \begin{tabularx}{0.95\linewidth}{lCCC}
        \toprule
        Dataset & PGexplainer & SubgraphX & Ours  \\
        \midrule
        BA-shapes &  0.606 & 0.364 & \textbf{0.105}  \\
        Syn-SST &  2.884 & 2.936 & \textbf{2.231}\\
        \bottomrule
    \end{tabularx}
    \vskip -1em
    \label{tab:syn}
\end{table}
\subsection{Classification and Explanation Quality}
To answer \textbf{RQ1}, we compare {\method} with the baselines in terms of prediction and quality of explanations with prototypes on both node classification and graph classification datasets. 

\subsubsection{Results on Real-World Graphs}
To demonstrate the effectiveness of our proposed {\method} in giving accurate predictions, we compare with state-of-the-art GNNs and self-explainable GNNs on real-world datasets of node classification and graph classification. We report the average results along with the standard deviations in Table~\ref{tab:pred}. Note that SE-GNN is particularly designed for node classification, so experiments on graph classification datasets are not applicable for SE-GNN.  From the Table~\ref{tab:pred}, we can observe that:
\begin{itemize}[leftmargin=*]
    \item Our {\method} outperforms GCN and GIN on various datasets, and achieves comparable results with SE-GNN that adopts contrastive learning. This is because the self-supervision of graph reconstruction and supervision from the labeled dataset are leveraged in {\method} for the encoder training and prototype generation, resulting better prediction performance;
    \item Compared with ProtGNN which also conducts self-explainable predictions with prototypes, the proposed {\method} gives better prediction results especially on node classification datasets. This demonstrates that {\method} can consistently learn high-quality prototype graphs for accurate predictions on different tasks; while simply selecting subgraphs from the figures as prototype graphs might result in sub-optimal prototypes. 
\end{itemize}

We further quantitatively evaluate the quality of the learned prototypes for explanations on real-world datasets with two evaluation metrics. \textit{First}, since prototype graphs should capture representative patterns of the corresponding class, the prediction confidence scores on the prototype graphs are expected to be high. Therefore, we evaluate the confidence score of the prototype graph using a trained GNN to assess the quality. \textit{Second},  high-quality prototype graphs can tightly cluster the instances in the datasets. In addition, clusters corresponding to different prototype graphs should be well separated. Therefore, we use the prototypes as centers and assign each node/graph to the nearest prototype. 
Then, we conduct Silhouette analysis~\cite{rousseeuw1987silhouettes} to evaluate the tightness and separateness of the clusters. The Silhouette score ranges from -1 to 1. A higher Silhouette score means tighter clusters and larger distance between clusters, which indicates more representative and diverse prototypes. The silhouette score is computed based on the embedding space learned by a graph autoencoder. For post-hoc explainers,  a GCN achieves the performance in Table~\ref{tab:pred} is set as the pretrained model.
The results in terms of Confidence Score and Silhoette score are given in Table~\ref{tab:proto}. From the table, we observe that:
\begin{itemize}[leftmargin=*]
    \item The confidence score and Silhouette score of prototype graphs learned by our {\method} are much larger than prototypes obtained from baseline GNN explainers. This demonstrates that our method can learn representative and diverse prototype graphs.
    \item Compared with ProtGNN which adopts Monte Carlo tree search to find subgraphs from raw graphs as prototypes, the prototypes generated by our {\method} are significantly better according to the quality scores. This is because the prototype generator in {\method} can learn prototypical patterns for classification with the deployed self-supervision and classification loss. 
\end{itemize}

\subsubsection{Results on Synthetic Datasets} To quantitatively evaluate the class-level explanations, we conduct experiments on BA-Shapes and Syn-SST which provide ground-truth prototypes. Following previous work~\cite{bai2019simgnn} in graph similarity measuring, we first pretrain a graph autoencoder and evaluate the distance of learned prototypes and the ground-truth prototypes in the embedding space. We compare our method with the most competitive methods in Table~\ref{tab:proto}. The results are shown in Table~\ref{tab:syn}. We can find that the learned prototype graphs from {\method} match much better with the ground-truth prototype graphs than the baseline prototypes. 

\subsubsection{Visualization} 
\begin{figure}[t]
    \small
    \begin{center}
    \begin{tabular}{cccc}
    \textbf{BA-shapes}: \vspace{-0.8em} &
    \\
    \includegraphics[width=0.085\linewidth,valign=T]{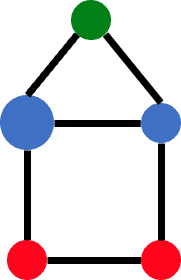}
    & \includegraphics[width=0.23\linewidth,valign=T]{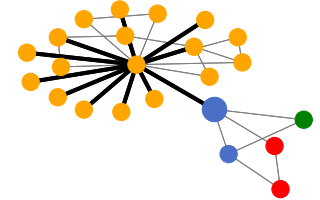}&
    \includegraphics[width=0.23\linewidth,valign=T]{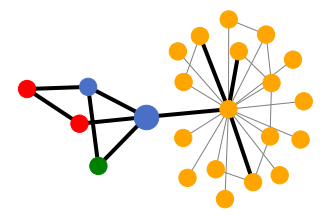}&
    \includegraphics[width=0.23\linewidth,valign=T]{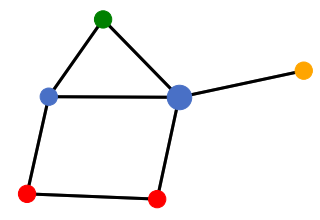} \vspace{0.3em}\\
    
    \textbf{MUTAG}: \vspace{-1em} &
    \\
    \includegraphics[width=0.2\linewidth,valign=T]{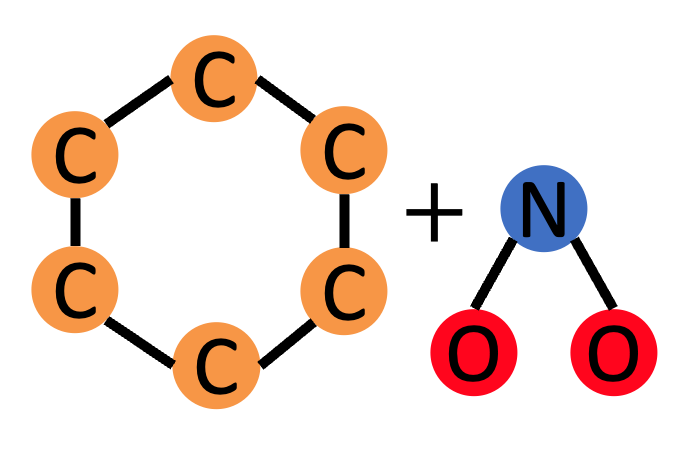}& 
    \includegraphics[width=0.25\linewidth,valign=T]{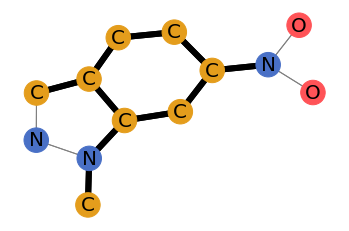}&
    \includegraphics[width=0.25\linewidth,valign=T]{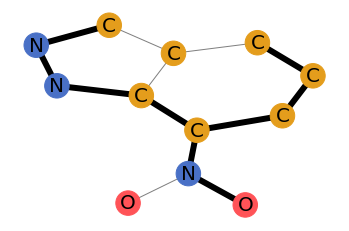}&
    \includegraphics[width=0.25\linewidth,valign=T]{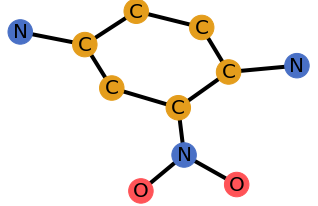}\\
    Ground Truth & PGExplianer & SubgraphX & Our {\method} \\ 
    \end{tabular}
    \end{center}
    \vskip -1.5em
    \caption{Learned prototypes on BA-Shapes and MUTAG.}
    \vskip -2em
    \label{fig:vis}
\end{figure}

To qualitatively evaluate the prototype-based explanations, we also visualize the obtained prototypes of our {\method} and baselines on both synthetic dataset BA-Shapes and real-world dataset MUTAG. The results are shown in Fig.~\ref{fig:vis}. For {\method}, we directly present the learned prototypes of the representative class. 
As for PGExplianer and SubgraphX, the prototypes are extracted crucial subgraph of the center graph. Thus, they are represented by the bold black edges of the complete center graph in Fig.~\ref{fig:vis}.
For the prototypes on BA-Shapes, we can find that our {\method} manage to learn prototype graphs that are consistent with the ground-truth of ``house'' motif. And {\method} also manage to abstract the carbon ring with $NO_2$ chemical group as the prototype for the real-world MUTAG dataset. On the other hand, the baseline explainers may cover some important links, but fail to learn a representative high-quality prototype graph for explanations.

\setlength{\tabcolsep}{1pt}

\subsection{Impacts of the Number of Prototypes}
\begin{figure}[t]
    \small
    \centering
    \begin{subfigure}{0.49\linewidth}
        \includegraphics[width=0.98\linewidth]{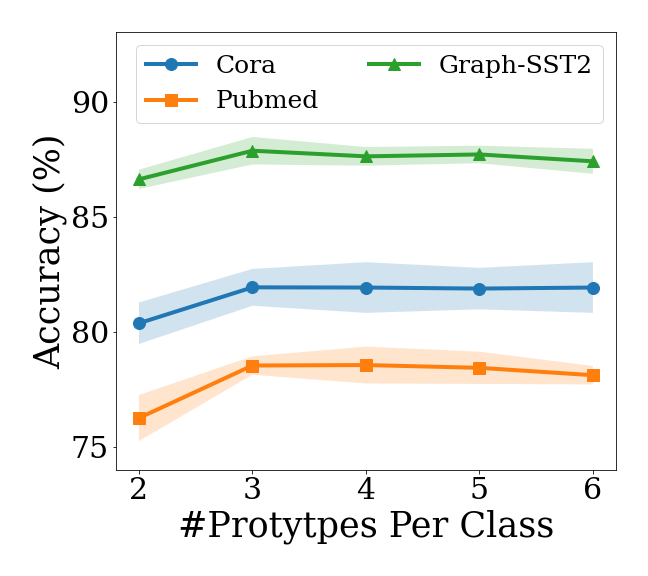}
        \vskip -1em
        \caption{Prediction Accuracy}
    \end{subfigure}
    \begin{subfigure}{0.49\linewidth}
        \includegraphics[width=0.98\linewidth]{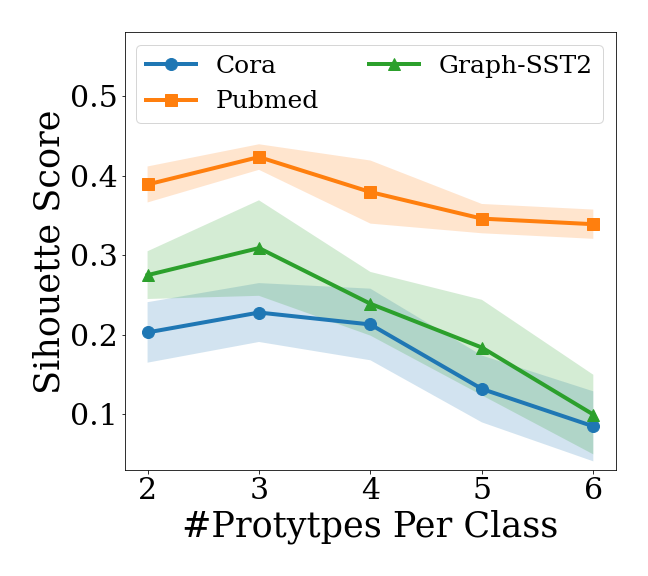}
        \vskip -1em
        \caption{Prototype Quality}
    \end{subfigure}
    \vskip -1.5em
    \caption{The impacts of number of prototypes per class.}
    \vskip -1.8em
    \label{fig:nclu}
\end{figure}

One natural question for prototype-based explanations is how to determine the size of prototypes. Thus, 
to answer \textbf{RQ2}, we vary the number of prototypes per class as $\{2, 3, \dots, 6\}$ to investigate its impacts to {\method}. The other hyperparameters are set as the description in Section \ref{sec:imple}.
The impacts are assessed from the aspects of prediction accuracy and the quality of learned prototypes. Here, the Silhouette score, which can measure the representativeness and diversity of the prototype graphs, are used for prototype quality evaluation. We only report the results on Cora, Pubmed and Graph-SST2 in Figure~\ref{fig:nclu} as we have similar observations on other datasets. From Figure~\ref{fig:nclu}, we observe that with the increase of prototype size, the classification performance will firstly increase then maintain similar results. For the Silhouette score, it will firstly increase then decrease. 
This is because when the number of prototypes is too small, the prototypes cannot represent all the instances in the dataset, leading to poor performance in prediction. When the prototype size is overly large, we will obtain multiple similar prototypes for one ground-truth prototype. In this situation, the prediction performance will not increase; while silhouette score which also measures the diversity of the prototypes will decrease. The above observations also pave us a way of selecting optimal prototype size in {\method}.

\subsection{Ablation Study}
\begin{figure}[t]
    \small
    \centering
    \begin{subfigure}{0.49\linewidth}
        \includegraphics[width=0.98\linewidth]{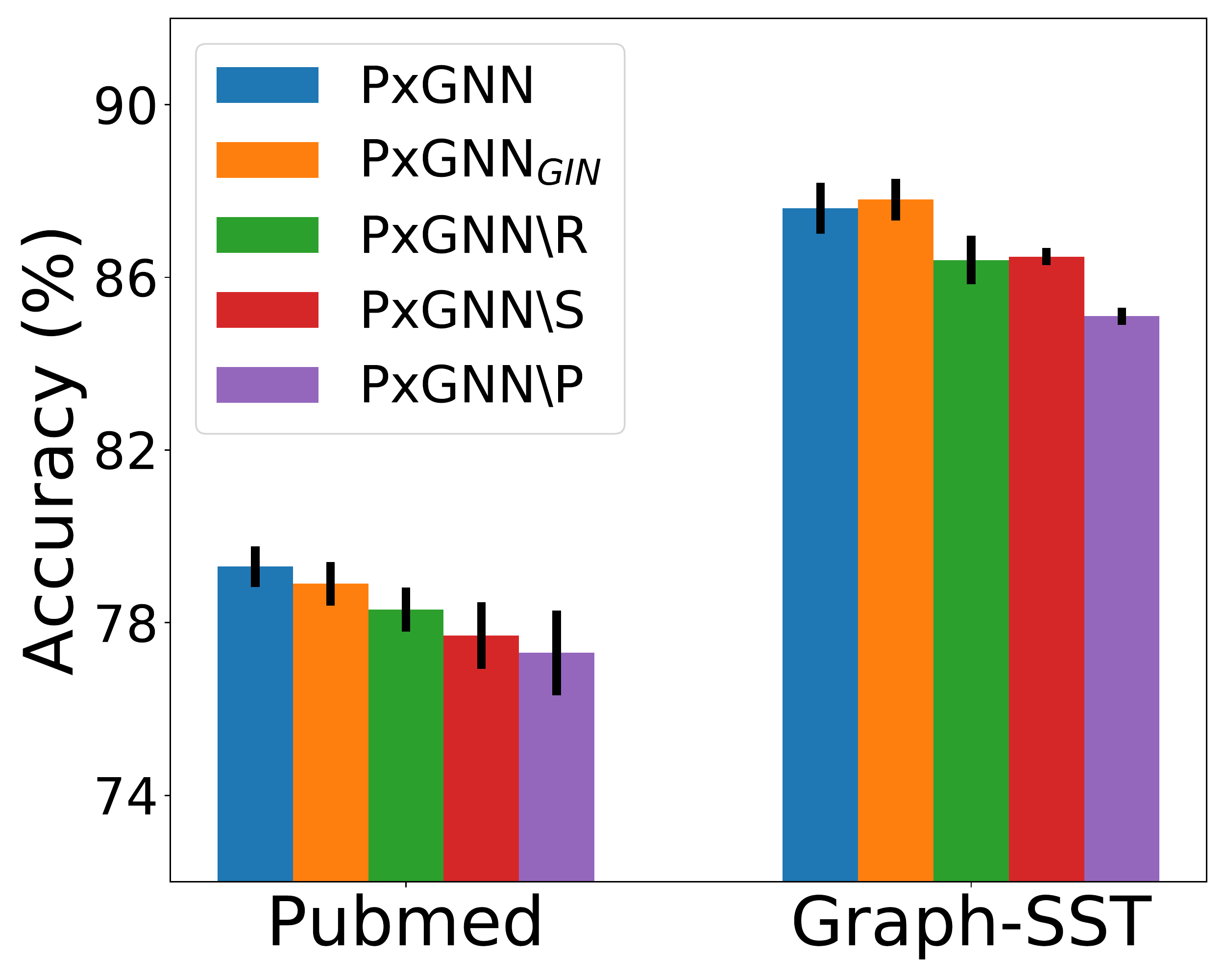}
        \vskip -1em
        \caption{Prediction Accuracy}
    \end{subfigure}
    \begin{subfigure}{0.49\linewidth}
        \includegraphics[width=0.98\linewidth]{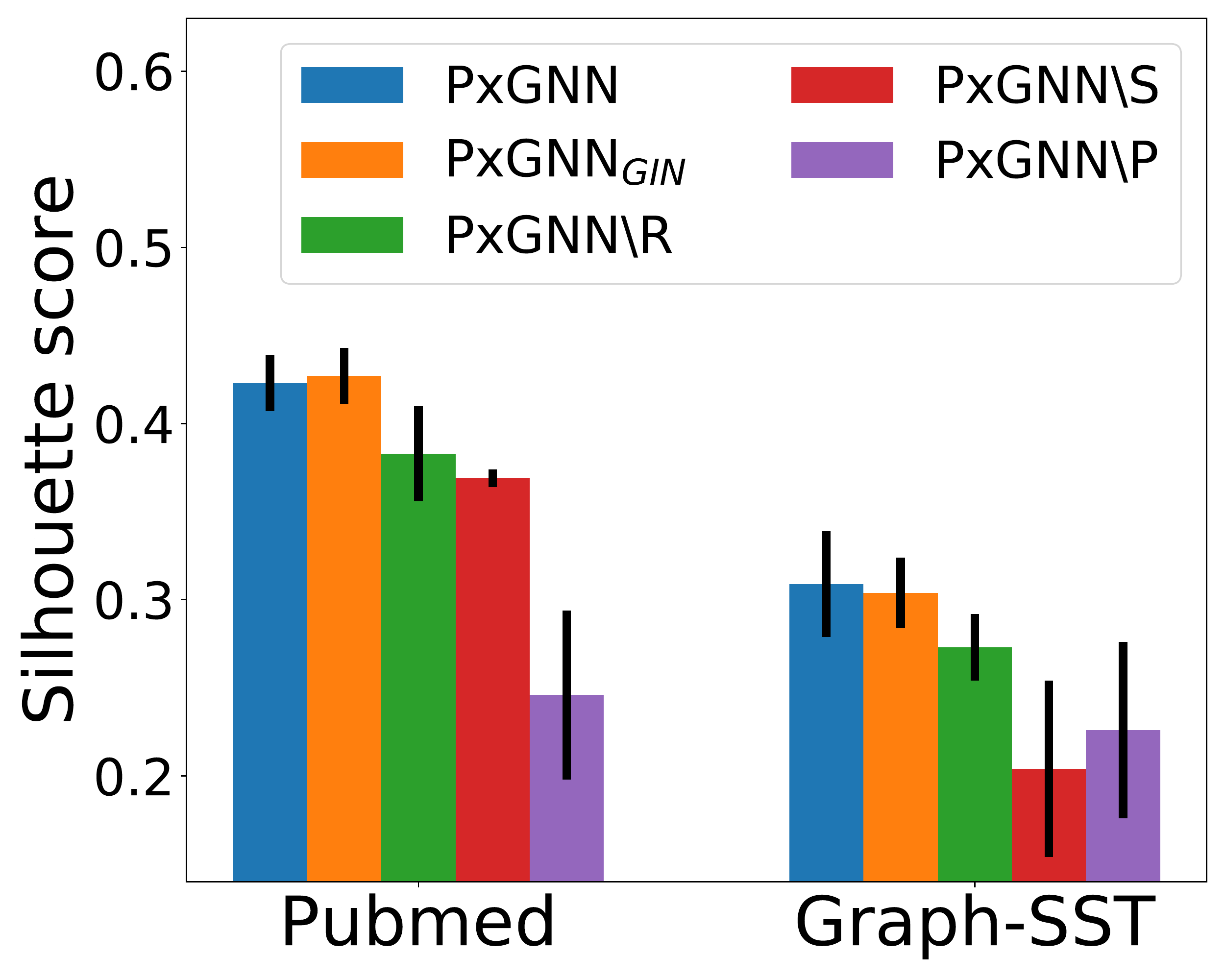}
        \vskip -1em
        \caption{Prototype Quality}
    \end{subfigure}
    \vskip -1.5em
    \caption{Ablation Study on Pubmed and Graph-SST2.}
    \vskip -1.5em
    \label{fig:abla}
\end{figure}

To answer \textbf{RQ3}, we conduct ablation studies to explore the flexibility of our proposed {\method} and the effectiveness of the self-supervision loss and constraint term in prototype generation.  To show that {\method} is flexible to various GNN bacbones, we replace the GCN-based encoder to a GIN model, denoted as {\method$_{GIN}$}. To demonstrate the importance of the self-supervision of graph reconstruction in prototype generator training, we train a variant {\method}$\backslash$S by setting $\alpha$ as 0. In {\method}, we utilize the initialization prototype graphs to constrain the prototype embeddings in the desired latent space to ensure the quality of generated prototypes. 
To show the effects of this constraint term, we set $\beta$ to 0 and obtain a variant named as {\method}$\backslash$R. Finally, to verify the necessity of updating the prototype graphs with the prototype generator, we train a variant {\method}$\backslash$P which fixes the prototype graphs after the initialization. The hyperparameters of these variants are also tuned on the validation set.
The average results on Pubmed and Graph-SST are presented in Figure~\ref{fig:abla}. We can observe that:
\begin{itemize}[leftmargin=*]
    \item {\method$_{GIN}$} achieves similar results with {\method} for both classification accuracy and prototype quality, which indicates the flexibility of the proposed {\method}; 
    \item The accuracy and prototype quality of {\method}$\backslash$S and {\method}$\backslash$R are significantly worse than {\method}. This demonstrates that the self-supervision on prototype generator and the constraint on prototype embeddings are helpful for learning high-quality prototypes for the prediction and explanation;
    \item {\method} outperforms {\method}$\backslash$P by a large margin, which proves the importance of updating the prototype graphs with generator to better capture the key patterns of each class. 
\end{itemize}

\subsection{Parameter Sensitivity Analysis}

\begin{figure}[t]
    \small
    \centering
    \begin{subfigure}{0.49\linewidth}
        \centering
        \includegraphics[width=0.98\linewidth]{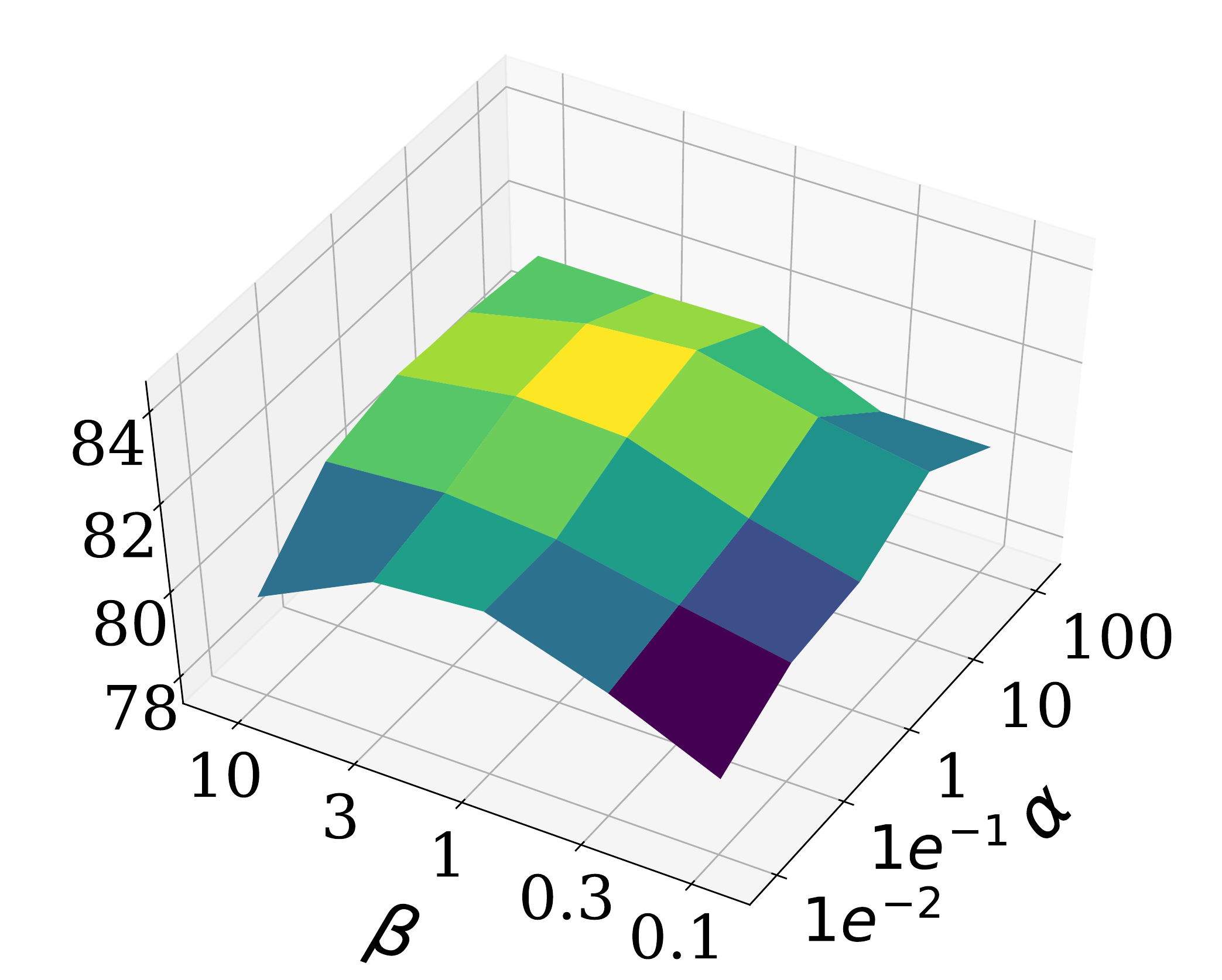}
        \vskip -0.5em
        \caption{Accuracy (\%)}
    \end{subfigure}
    \begin{subfigure}{0.49\linewidth}
        \centering
        \includegraphics[width=0.98\linewidth]{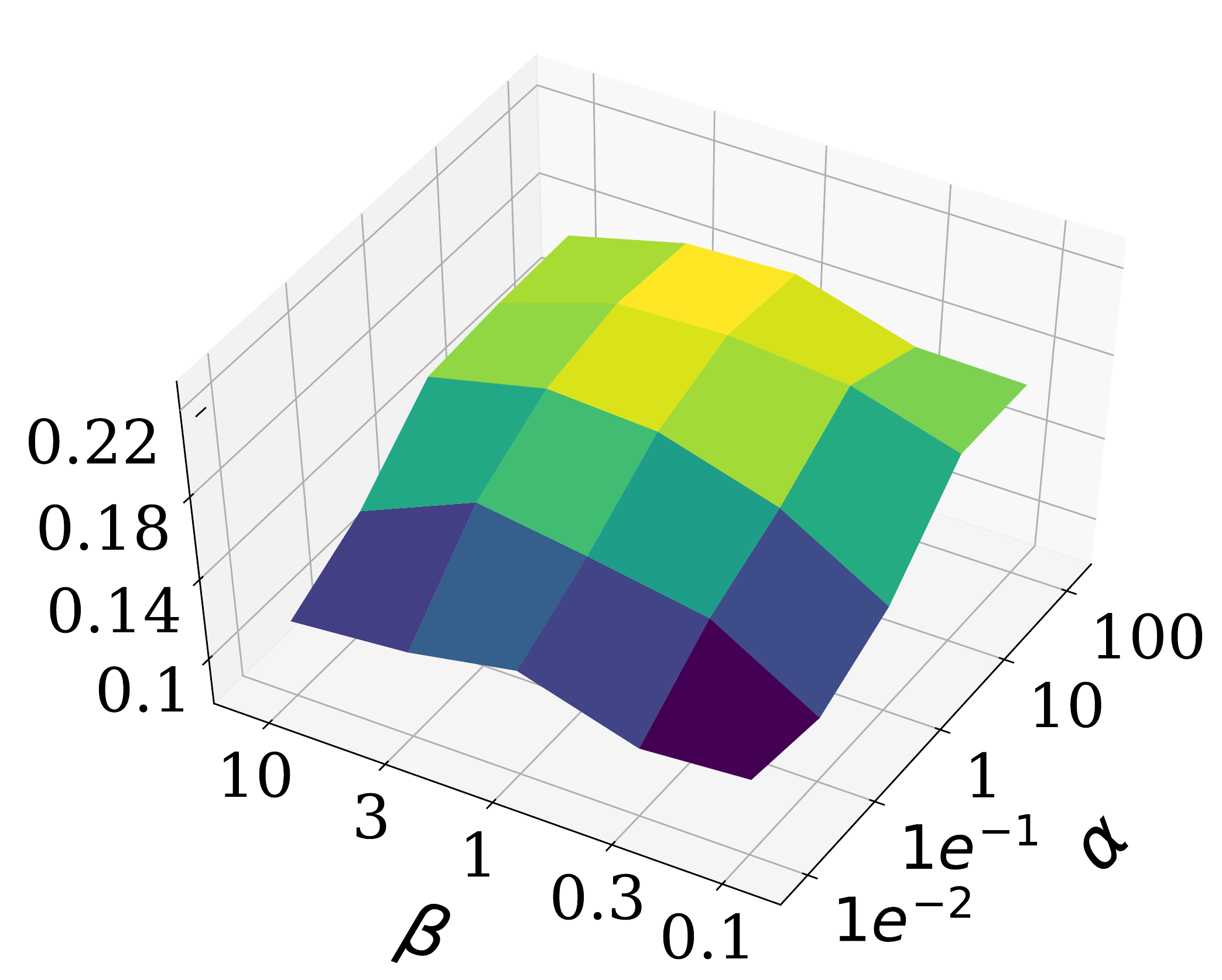}
        \vskip -0.5em
        \caption{ Silhouette Score}
    \end{subfigure}
    \vskip -1.5em
    \caption{Parameter sensitivity analysis on Cora.}
    \vskip -1.5em
    \label{fig:para}
\end{figure}

In this subsection, we study how the hyperparameters $\alpha$ and $\beta$ will affect {\method}. $\alpha$ controls the contribution of self-supervision on encoder and prototype generator. And $\beta$ controls the regularization applied on the learnable prototype embeddings. We alter the values of $\alpha$ and $\beta$ as $\{100,10,1,0.1,0.01\}$ and $\{10,3,1,0.3,0.1\}$, respectively. The setting of other hyperparameters are the same as the description in Section~\ref{sec:imple}. The results are presented in Figure~\ref{fig:para}. From the figure, we observe that: (\textbf{i}) when $\alpha$ is small, both the classification accuracy and prototype quality are poor.  This is because little contribution of self-supervision would lead to a weak prototype generator. And a too large $\alpha$ will also lead to the decrease of classification accuracy because the reconstruction loss dominate the whole loss function; and (\textbf{ii}) with the increase of $\alpha$, the performance in both aspects will firstly increase then decrease. When $\beta$ is overly small, the updated  prototype embedding may not perfectly follow the latent space for prototype generation, leading to relatively poor prototypes. However, a too large $\beta$ will largely restrict the updates of prototype embeddings.  Combining the two figures, we can see that when $\alpha \in [1,10]$ and $\beta \in [1,3]$, {\method} can give accurate predictions and obtain high-quality prototypes for explanations.

\section{Conclusion and Future Work}
In this paper, we study a novel problem of learning a prototype-based self-explainable GNN. We develop a new framework {\method}, which adopts a prototype generator to learn representative and realistic prototype graphs for accurate prediction and explanations. The self-supervision of graph reconstruction and the supervision from labeled instances is applied to facilitate the learning of prototypes and the performance of classification. Extensive experiments on real-world and synthetic datasets demonstrate the effectiveness of our {\method} in self-explainable classification on nodes and graphs. Further experiments are conducted to explore the optimal prototype size, the contributions of each component in {\method}, and the hyperparameter sensitivity. There are some interesting directions which require further investigation. One direction is to extend {\method} to heterogeneous graphs. It is also promising to explore how the prototype-based explanations can be utilized to achieve robustness and fairness.

\bibliographystyle{ACM-Reference-Format}
\bibliography{acmart}

\end{document}